\documentclass{article}

\usepackage{arxiv}

\usepackage[utf8]{inputenc} 
\usepackage[T1]{fontenc}    
\usepackage{hyperref}       
\usepackage{url}            
\usepackage{booktabs}       
\usepackage{amsfonts}       
\usepackage{nicefrac}       
\usepackage{microtype}      
\usepackage{graphicx}
\usepackage{doi}
\usepackage{times} 
\usepackage{amsmath} 
\usepackage{multirow}
\usepackage{textcomp}
\usepackage{pgfplots}
\usepackage{mathrsfs} 
\usepackage{amssymb}  
\usepackage[noadjust]{cite}
\usepackage{footnote}
\makesavenoteenv{tabular}
\makesavenoteenv{table}
\makeatletter

\title{Cross-Quality LFW: A Database for Analyzing Cross-Resolution Image Face Recognition in Unconstrained Environments}

\author{\href{https://orcid.org/0000-0002-0503-4600}{\includegraphics[scale=0.06]{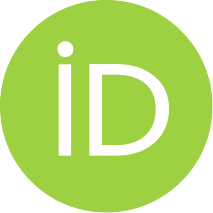}\hspace{1mm}Martin~Knoche}~~~~Stefan~H\"orman~~~~\href{https://orcid.org/0000-0003-1096-1596}{\includegraphics[scale=0.06]{orcid.pdf}\hspace{1mm}Gerhard ~Rigoll}\\
	Chair of Human-Machine Communication\\
	Technical University\\
	Munich, Germany \\
	\texttt{\href{Martin.Knoche@tum.de}{Martin.Knoche@tum.de}} \\
}

\renewcommand{\headeright}{A Preprint}
\renewcommand{\undertitle}{A Preprint  \\ 
Published in International Conference on Automatic Face and Gesture Recognition 2021 \\ 
DOI: \url{https://doi.org/10.1109/FG52635.2021.9666960}}
\renewcommand{\shorttitle}{XQLFW database}

\hypersetup{
pdftitle={xqlfw},
pdfsubject={},
pdfauthor={Martin~Knoche, Stefan~H\"orman, Gerhard~Rigoll},
pdfkeywords={recognition, resolution, cross, face, identification, verification, dataset, database, benchmark, evaluation, protocoll, image},
}

\begin{document}
\maketitle

\begin{abstract}
Real-world face recognition applications often deal with suboptimal image quality or resolution due to different capturing conditions such as various subject-to-camera distances, poor camera settings, or motion blur. This characteristic has an unignorable effect on performance. Recent cross-resolution face recognition approaches used simple, arbitrary, and unrealistic down- and up-scaling techniques to measure robustness against real-world edge-cases in image quality. Thus, we propose a new standardized benchmark dataset and evaluation protocol derived from the famous Labeled Faces in the Wild (LFW). In contrast to previous derivatives, which focus on pose, age, similarity, and adversarial attacks, our Cross-Quality Labeled Faces in the Wild (XQLFW) maximizes the quality difference. It contains only more realistic synthetically degraded images when necessary. Our proposed dataset is then used to further investigate the influence of image quality on several state-of-the-art approaches. With XQLFW, we show that these models perform differently in cross-quality cases, and hence, the generalizing capability is not accurately predicted by their performance on LFW. Additionally, we report baseline accuracy with recent deep learning models explicitly trained for cross-resolution applications and evaluate the susceptibility to image quality. To encourage further research in cross-resolution face recognition and incite the assessment of image quality robustness, we publish the database and code for evaluation.\footnote{Code, dataset and evaluation protocol available at \url{https://martlgap.github.io/xqlfw}}
\end{abstract}

\section{INTRODUCTION}

Current state-of-the-art face recognition systems~\cite{deng2019arcface, meng2021magface, zhong2021face} show superior performance on several standard face recognition benchmarks in unconstrained environments (e.g., MegaFace~\cite{MegaFace}, IJB-A~\cite{IJBA}, or LFW~\cite{LFW}), almost reaching saturation levels on LFW. But do these results generalize to more challenging scenarios or edge-cases? Multiple benchmarks focus on specific properties like age~\cite{AgeDB, CALFW} or pose~\cite{CPLFW, CFP} to enhance the difficulty, which results in a substantial drop in performance. Other works on occlusions~\cite{PartialLFW} or transferable adversarial attacks~\cite{TALFW} also report decreasing performances with modified databases. 

In real-world face recognition applications, the quality and resolution of the examined images vary due to different camera settings or the subject-to-camera distance. Comparing two faces, comprising a different image resolution or quality, is often referred to as cross-resolution problem. This inequality of image resolution substantially affects the performance, and hence, several methods studied cross-resolution face verification~\cite{knoche2021image, massoli2020cross, 8303213, zangeneh2020low, khazaie2020ipu, talreja2019attribute, mudunuri2018genlr, ge2018low, 7550087}. To evaluate cross-resolution face recognition systems, they simulate a lower image resolution by down- and up-sampling with bicubic or bilinear kernels. However, several studies~\cite{ji2020real, zhang2019deep, zhou2019kernel, bell2019blind} on image super-resolution showed that real low-resolution images differ from synthetically generated images. Moreover, sampling kernels vary across software packages and make a fair comparison impossible. This circumstance motivates us to dig deeper into the LFW database and analyze it according to image resolution and quality.

\begin{figure}[t!]
  \centering
  \includegraphics[width=\columnwidth]{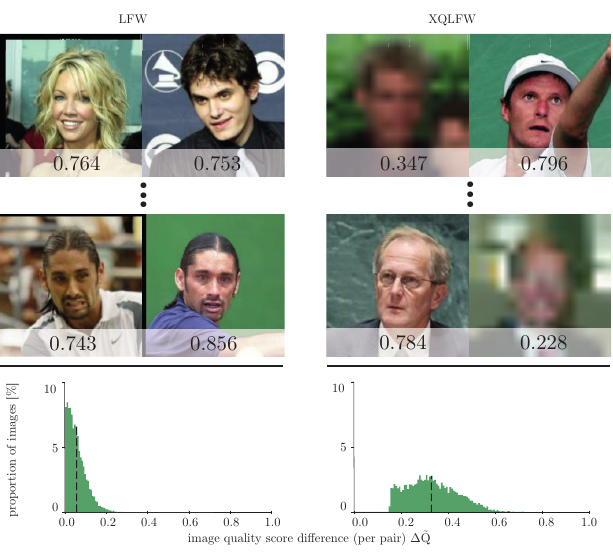}
  \caption{Comparison between LFW and our proposed XQLFW dataset with two example image pairs, image quality scores, and the distribution of the absolute difference in image quality per pair.}
  \vspace{-0.4cm}
  \label{fig:LFWvsXQLFW}
\end{figure}

Modern face recognition systems often rely on a fixed input image resolution of $112\times112\,\text{px}$ determined by the resolution of training images. But the pixel dimensions do not necessarily correspond to the real resolution of the image. LFW, for example, contains loosely cropped images with $250\times250\,\text{px}$ resolution, with a facial region covering $\approx 112\times112\,\text{px}$. We cannot be sure that all images are captured with exactly that resolution. Since images are crawled from the web, they are very likely down- or up-sampled beforehand. Besides characteristic camera motion blur or bad lighting conditions, the lower original image resolution constitutes a reason for encountering images with inferior quality in the LFW database. However, the inherent quality difference within pairs of the LFW database is tiny. \autoref{fig:LFWvsXQLFW} shows two example pairs of LFW and the distribution of image quality score differences per pair, demonstrating a small variance of image quality within the database. 

To enlarge the quality variance and thus to test the robustness of face recognition systems against image quality, we use a more realistic method to synthetically deteriorate images and create a modified Cross-Quality LFW dataset (XQLFW). \autoref{fig:LFWvsXQLFW} depicts two example pairs of our proposed XQLFW, demonstrating a more considerable quality difference per pair. The distribution of quality score differences per pair of our XQLFW protocol compared to LFW in \autoref{fig:LFWvsXQLFW} is shifted to the right, which indicates a wider variety of image quality in our proposed evaluation protocol.

Our main contributions can be summarized as follows:
\begin{itemize}
    \item We continue research on the LFW database and show that image quality variation inside the evaluation protocol is tiny.
    \item We establish a more challenging and realistic database to evaluate the robustness of face recognition systems towards cross-resolution image quality.
    \item We maintain the dataset size, image ensemble, and face verification protocol rules of LFW and thereby ensure disjoint identities in training and testing datasets.
    \item We report and analyze the robustness of face recognition performance for several state-of-the-art approaches on our novel XQLFW evaluation protocol and reveal large discrepancies in the generalization performance of several state-of-the-art face recognition approaches.
\end{itemize}

\section{RELATED WORK}

\subsection{Datasets}
Most of the publicly available databases concentrate exclusively on rather high-resolution images (e.g., LFW~\cite{LFW}, IJB-A~\cite{IJBA}, IJB-B~\cite{IJBB}, MegaFace~\cite{MegaFace}, AgeDB~\cite{AgeDB}, CFP~\cite{CFP}) or rather low-resolution images (e.g. TinyFaces~\cite{TinyFace}). SCFace~\cite{SCFace}, for example, combines high- and low-resolutions but contains only $130$ subjects, which makes it not suitable for performance evaluation of cross-resolution face recognition due to its poor generalization. 

To overcome the saturating performance on the LFW dataset, other LFW derivatives were created in recent years. For example, Zheng et al.~\cite{CALFW, CPLFW} focused on significant age and pose variance within the evaluation protocol and provided the cross-age and cross-pose LFW dataset (CALFW, CPLFW). While Zhong et al.~\cite{TALFW} investigated the vulnerability of face recognition systems against transferable adversarial attacks and proposed a novel Transferable Adversarial LFW dataset (TALFW), Deng et al.~\cite{SLLFW} analyzes the robustness against similar looking faces and propose a new challenging evaluation protocol (SLLFW). 

Facial occlusions are studied in the following works: The authors of~\cite{PartialLFW} analyzed the impact of occlusions and proposed the PartialLFW evaluation dataset, which contains face images with synthetically generated  occlusions on different facial landmarks. Eyeglass-robustness of face recognition systems was analyzed by Guo et al.~\cite{MeGlass}. Recently, in~\cite{wang2020masked} and~\cite{montero2021boosting} LFW is extended with synthetically added face masks.

In contrast, our XQLFW evaluation protocol aims to show how well face recognition performs under realistic edge-case scenarios in the scope of image quality.

\subsection{Image Quality Metrics}

Reference-based image quality assessment approaches compare the image quality between two or more images. In contrast, no-reference-based methods focus on a single image and report an independent, absolute score, representing the quality of the image. In this work, we use no-reference-based metrics to assess each image's quality independently. Kamble et al.~\cite{surveyimquality} presented an exhaustive enumeration of no-reference-based approaches in their work. 

We distinguish between non-face-specific metrics (e.g., BRISQUE~\cite{brisque}, sharpness~\cite{kumar2012sharpness}) and face-specific metrics in the following: In~\cite{dfqa}, the authors presented a two-stream convolutional neural network that quickly and accurately predicts the face quality score. Another approach was suggested by Khryashchev et al.~\cite{KhryashchevFaceImg}. They propose a novel metric-based image quality assessment using resolution, sharpness, symmetry, blur, and face landmarks. Recently, Terh\"orst et al.~\cite{serfiq} studied the unsupervised estimation of face image quality based on stochastic embedding robustness (SER-FIQ).

\subsection{Cross-Resolution Face Recognition}
Cross-resolution face recognition can be categorized into two groups: 1) transformation-based approaches~\cite{8303213, khazaie2020ipu, talreja2019attribute}, which first transform images into the same resolution or quality space and then apply face recognition. 2) non-transformation based approaches~\cite{mudunuri2018genlr, massoli2020cross, singh2018identity, knoche2021image}, which directly project facial features of different image resolutions/qualities into the same space. 

However, a fair performance comparison across different approaches is impeded by different evaluation protocols and image down-sampling methods for benchmarks. Thus, we deduct the need for a standardized cross-resolution database to more accurately gauge performance on realistic and challenging cross-quality images.

\section{CONSTRUCTING XQLFW}

\subsection{Image Quality Assessment}
\label{sec:imagequality}
First, we evaluate the BRISQUE~\cite{brisque} and SER-FIQ~\cite{serfiq} scores on every single image of the LFW database. To mitigate effects from the background around the face, we crop and align all images with MTCNN~\cite{mtcnn} as in~\cite{deng2019arcface}. While BRISQUE measures the visual quality of an image, SER-FIQ evaluates the quality of the face itself via facial feature assessment (i.e., occlusions or extreme head poses) result in a meaningless identity feature and thus  reflect poor quality). A correlation coefficient of $-0.021$ for the complete LFW database proves the independence of both metrics. For a combination of both, we first normalize each score ${Q}_m(\cdot),\,m \in \{{\text{BRISQUE, SER-FIQ}}\}$ for a given image $\boldsymbol{I}$ such that they both lie in the same value range $[0,1]$:
\begin{equation}
    \Tilde{Q}_m(\boldsymbol{I}) = \frac{\text{min}(\text{max}(q_{\text{min,  }m},Q_{m}(\boldsymbol{I})),q_{\text{max,} m})}{q_{\text{max,} m}-q_{\text{min,} m}}
\end{equation}
\begin{figure}[b!]
  \centering
  \includegraphics[width=\columnwidth]{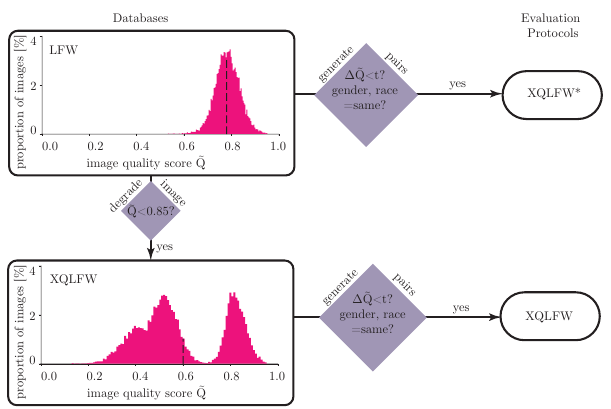}
  \caption{Pipeline for the construction of the proposed cross-quality labeled faces in the wild (XQLFW) evaluation protocol.}
  \label{fig:flowchart}
\end{figure}
For $\Tilde{Q}_{\text{BRISQUE}}(\cdot)$, we set $q_{\text{min, BRISQUE}}=0$ and $q_{\text{max, BRISQUE}}=100$, which correspond to the minimum and maximum possible values reported in~\cite{brisque}. During the normalization of $\Tilde{Q}_{\text{SER-FIQ}}(\cdot)$, we use minimum and maximum values of $0.78$ and $0.91$, to capture the full range of image quality. Then, we calculate the normalized combined image quality score $\Tilde{Q}(\cdot)$ for a given image $\boldsymbol{I}$ using both normalized scores as follows:
\begin{equation}
    \Tilde{Q}(\boldsymbol{I}) = \frac{1 - \Tilde{Q}_{\text{BRISQUE}}(\boldsymbol{I}) + \Tilde{Q}_{\text{SER-FIQ}}(\boldsymbol{I})}{2}
\end{equation}

\subsection{Evaluation Protocol Construction Details}
The Labeled Faces in the Wild database (LFW) contains $13233$ images from $5749$ unique identities. The number of images per identity varies from $1$ up to $530$. The LFW View-2 evaluation protocol defines $3000$ image pairs with the same identity (genuine) and $3000$ image pairs with different identity (imposter). 

To construct the XQLFW evaluation protocol, we follow the same procedure as Huang et al.~\cite{LFW} proposed for the LFW View-2 evaluation protocol. Genuine pairs are formed iteratively: First, we randomly pick one identity from all identities with at least two images. Two different images were selected at random from this given identity and added to the evaluation protocol if that specific pair was not already added previously. The whole process is repeated until 3000 pairs are found.

Imposter pairs are formed iteratively as follows: First, we randomly pick two identities out of all identities. If this specific combination of identities is already present in the protocol, we repeat this step. From each identity, one image is then selected at random. Similar to~\cite{CPLFW} and~\cite{CALFW}, gender and race are forced to be equal by using the attributes provided by~\cite{LFW}. This process is also repeated until 3000 pairs are generated.

\subsection{Synthetic Image Quality Deterioration}
To further increase the per pair image quality score difference $\Delta \Tilde{Q}$ within the evaluation protocol, we introduce a threshold $t$ for a minimum score difference and apply synthetic image deterioration. The process is described as follows: We loop over each identity in the database and synthetically deteriorate images if: 1) $\Delta \Tilde{Q}$ is below a quality threshold of $0.85$ and 2) the number of deteriorated images within an identity does not exceed half the number of images of that identity. This procedure assures that a certain amount of high-quality images remains for each identity. 

Several works~\cite{bell2019blind,ji2020real, zhang2019deep, zhou2019kernel} argue that simple down- and up-sampling images with, e.g., bicubic or bilinear kernels, is not sufficiently reflecting real low image resolution. In contrast to previous works, we therefore use the method from Bell et al.~\cite{bell2019blind} to blur and then sub-sample each image with a different, randomly generated $21\times21\,\text{px}$ Gaussian an-isotropic kernel. Scale factors are randomly chosen from the following list $\{3, 4, 5, 6, 7, 8, 10, 12, 14, 16\}$. Additionally, we specifically apply deterioration to rather low-quality images only. Experiments demonstrate that this synthetic image resolution reduction highly correlates with our combined image quality score. 

As depicted in~\autoref{fig:flowchart}, we follow this construction protocol and generate two evaluation protocols using a threshold $t = 0.15$: 1) XQLFW*, which is developed from the original LFW database and 2) XQLFW, which is developed from the synthetically deteriorated XQLFW database. \autoref{fig:flowchart} additionally illustrates the different distributions of image quality scores within both source databases.

\begin{figure}[b!]
  \centering
  \includegraphics[width=\columnwidth]{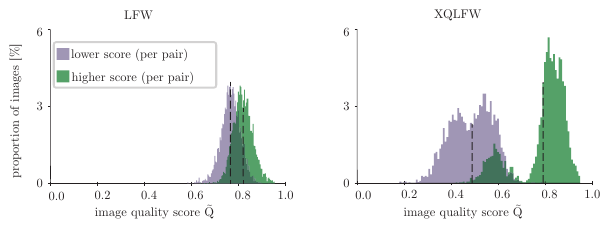}
  \caption{Combined image quality score distribution $\Tilde{Q}$ for LFW and XQLFW with the lower (grey) and higher (green) score of every image pair.}
  \label{fig:hist_pairs}
\end{figure}

\subsection{Comparison of LFW and XQLFW}

\begin{table}[t!]
  \centering
  \caption{Comparison of the mean image quality difference $\varnothing \Delta \Tilde{Q}$ and number of unique identities and images of several databases.}
    \begin{tabular}{lcccccc}
    \toprule
          & LFW & XQLFW* & XQLFW & CPLFW & CALFW & SLLFW \\
    \cmidrule{2-7}    
    $\varnothing \Delta \Tilde{Q}$ & 0.056  & 0.177  & 0.327  & 0.078 & 0.046 & 0.054 \\
    identities & 4281  & 2450  & 3743  & 2296 & 2997 & 2810 \\
    images & 7701  & 4395  & 7263  & 5984 & 7167 & 6091 \\
    \bottomrule
    \vspace{0.3cm}
    \end{tabular}%
  \label{tab:datasetstats}%
\end{table}%

The LFW database mainly contains images with a score $\Tilde{Q}$ in the range of $0.7$ to $0.9$. In contrast, XQLFW principally consists of two groups of images with $\Tilde{Q}$ in the range of $0.7$ to $0.9$ corresponding to the non-degraded images from the LFW, and additionally, $0.25$ to $0.65$. This database characteristic enables imposter and genuine image pairs with an extensive image quality score difference. We report an average $\Delta \Tilde{Q}$ of $0.177$ for XQLFW* and $0.327$ for XQLFW, compared to the relatively low average $\Delta \Tilde{Q}$ of $0.056$ for LFW (c.f.~\autoref{tab:datasetstats}). The comparatively large average $\Delta \Tilde{Q}$ score of $0.078$ in CPLFW also reveals the susceptibility of the SER-FIQ metric against extreme head pose variations. 

In \autoref{fig:hist_pairs}, we depict the image quality distribution for LFW and XQLFW. To highlight the quality score differences between both images of a pair, we employ different colors for the lower and higher image quality score of every image pair. In contrast to LFW, one can see the widening gap of scores in XQLFW, which implicates a significantly larger quality score difference (c.f. also \autoref{fig:LFWvsXQLFW}). By not strictly picking the image with higher $\Tilde{Q}$ exclusively but also from the degraded images and allowing to select a degraded image, we also include a relatively small amount of cross-resolution pairs with both images having rather low quality (c.f.~\autoref{fig:hist_pairs}). 

Moreover, we state the number of unique identities and images for the resulting evaluation protocols and compare them to LFW, CALFW, CPLFW, and SLLFW (c.f.~\autoref{tab:datasetstats}). Due to the relatively small image quality variations within the LFW database, the construction of XQLFW* leads to excessive use of particular identities and images. E.g., mainly the rare identities with large quality variation within the images are preferably chosen for genuine pairs. Consequently, the XQLFW* evaluation protocol contains only $2450$ identities and $4395$ unique images, thus lacking generality. CALFW, CPLFW, and SLLFW similarly have fewer unique identities and images compared to LFW. However, our proposed evaluation protocol (XQLFW), derived from the deteriorated database, contains $3743$ individual identities and $7263$ images, topmost among other LFW derivatives mentioned in \autoref{tab:datasetstats}.

\section{FACE RECOGNITION BENCHMARK}

\begin{table}[b!]
  \centering
  \caption{Face verification accuracy (\%) for several state-of-the-art approaches on LFW and our generated evaluation protocols. The absolute decrease with respect to LFW is shown in brackets.}
    \begin{tabular}{llll}
    \toprule
    Model & LFW & XQLFW* & XQLFW \\
    \midrule
    ArcFace~\cite{deng2019arcface}\footnote{We trained this network using the following re-implementation of ArcFace \url{https://github.com/peteryuX/arcface-tf2}} & $99.50$ & $99.13\,(-0.37)$ & $74.22\,(-25.28)$ \\
    MagFace~\cite{meng2021magface} & $99.63$ & $\mathbf{99.35}\,(-0.28)$ & $76.95\,(-22.68)$ \\
    FaceTransformer~\cite{zhong2021face} & $\mathbf{99.70}$ & $\mathbf{99.35}\,(-0.35)$ & $87.90\,(-11.80)$ \\
    BT-M~\cite{knoche2021image} & $99.30$ & $99.10\,(-0.20)$ & $83.60\,(-15.70)$ \\
    ST-M1~\cite{knoche2021image} & $97.30$ & $96.50\,(-0.80)$ & $\mathbf{90.97}\,(-6.33)$ \\
    ST-M2~\cite{knoche2021image} & $95.87$ & $94.77\,(-1.10)$ & $90.82\,(-5.05)$ \\
    \bottomrule
    \end{tabular}%
    \label{tab:results}%
\end{table}%

We benchmark the evaluation protocols XQLFW* and XQLFW with several state-of-the-art face recognition approaches (all using a cleaned version of MS1M~\cite{guo2016ms} for training). Table~\ref{tab:results} depicts our XQLFW, the original LFW, and the non-degraded variant XQLFW* with a threshold $t = 0.15$ and denotes face verification accuracy. The small decrease in performance for XQLFW* underlines the requirement of further deterioration of images to measure the susceptibility of face recognition systems to image quality. While the performance on XQLFW drops substantially for ArcFace~\cite{deng2019arcface} and  MagFace~\cite{meng2021magface}, the decrease of accuracy for BT-M, ST-M1, and ST-M2 from~\cite{knoche2021image} is moderate. This comparatively small worsening is reasonable due to the specific training methods, which aim for a resolution-robust network. However, the performance of ST-M1 and ST-M2 is considerably lower on LFW, which is a huge drawback.

Interestingly, the performance of the FaceTransformer approach~\cite{zhong2021face} is remarkably good on the XQLFW protocol. We conclude that the Transformer~\cite{vaswani2017attention} architecture, which is heavily used in speech recognition, applied in the FaceTransformer network, is less susceptible to image resolution or quality than classical CNN architectures.

\begin{figure}[t!]
  \centering
  \includegraphics[width=\columnwidth]{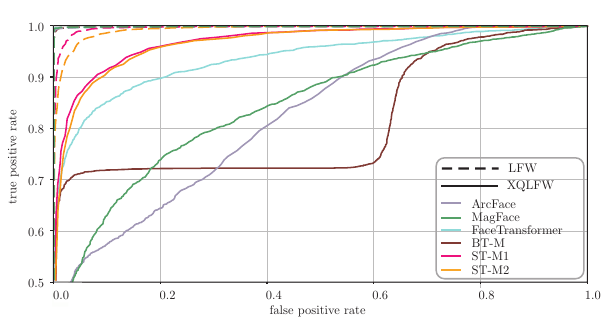}
  \caption{ROC curves of several state-of-the-art face recognition approaches on LFW and our proposed XQLFW evaluation protocol.}
  \vspace{0.5cm}
  \label{fig:roc}
\end{figure}

\autoref{fig:roc} shows the receiver operating characteristic curve for several state-of-the-art models on LFW and our proposed XQLFW evaluation protocol. Interestingly, the FaceTransformer approach outperforms all other models on XQLFW at very low false positive rates, whereas ST-M1 and ST-M2 clearly perform best at higher false positive rates. The BT-M model performs considerably better for about two-thirds of the database but struggles with the remaining third.

\section{CONCLUSIONS}

This paper introduces a novel face recognition benchmark protocol constructed from the well-known LFW database: Cross-Quality Labeled Faces in the Wild (XQLFW). This dataset focuses on significant image quality variations and thus, evaluates face recognition systems on their robustness against image quality or resolution. We first synthetically deteriorate a fraction of images from the original LFW database via blurring with random variations of Gaussian kernels to enhance the quality variation. We then randomly generate image pairs and construct our evaluation protocol while: 1) maintaining the characteristics of the original LFW evaluation protocol (View-2), hence, being easy-integrable, 2) keeping gender and race equality to be consistent with genuine pairs, and 3) using a considerable number of identities to preserve generality. A benchmark of several state-of-the-art approaches shows that superior face recognition performance on standard datasets like LFW is not necessarily correlated to the challenging and more realistic XQLFW. We conclude that our dataset provides new insights and helps to better understand and further develop real-world applicable face recognition systems.


{\small
\bibliographystyle{IEEEbib}
\bibliography{main}
}

\end{document}